\documentclass{openmoss}
\usepackage{amsmath,amsfonts,bm}

\def\eqref#1{equation~\ref{#1}}

\def\1{\bm{1}}

\DeclareMathAlphabet{\mathsfit}{\encodingdefault}{\sfdefault}{m}{sl}
\SetMathAlphabet{\mathsfit}{bold}{\encodingdefault}{\sfdefault}{bx}{n}

\usepackage{graphicx}
\usepackage{wrapfig}
\usepackage{needspace}
\usepackage{float} 
\usepackage[most]{tcolorbox}
\newtcolorbox{appendixprompt}[1]{
  enhanced,breakable,
  colback={rgb,255:red,250;green,250;blue,255},
  colframe={rgb,255:red,83;green,77;blue,158},
  colbacktitle={rgb,255:red,184;green,177;blue,255},
  coltitle=black,fonttitle=\bfseries,
  fontupper=\small,boxrule=0.8pt,arc=2pt,
  left=9pt,right=9pt,top=9pt,bottom=8pt,
  before skip=12pt,after skip=12pt,
  attach boxed title to top left={xshift=7pt,yshift=-2mm},
  boxed title style={boxrule=0pt,arc=2pt,left=5pt,right=5pt,
    top=2pt,bottom=2pt},
  title={#1},title after break={#1 (continued)}}
\newtcolorbox{agentmessage}[2]{
  enhanced,breakable,
  colback=white,colframe={rgb,255:red,35;green,101;blue,210},
  colbacktitle=white,coltitle={rgb,255:red,25;green,52;blue,91},
  fonttitle=\small\bfseries,fontupper=\small,
  frame hidden,borderline west={1.3pt}{0pt}{blue!70!black},
  boxrule=0pt,arc=0pt,left=8pt,right=6pt,top=4pt,bottom=4pt,
  before skip=7pt,after skip=7pt,
  title={#1\hfill\normalfont\footnotesize #2}}
\usepackage{colortbl}
\definecolor{appendixblue}{RGB}{233,244,252}
\usepackage{booktabs}
\usepackage{hyperref}
\hypersetup{pdftitle={Embodied Task Agents Can Perform Fine Manipulation through Tactile In-Context Learning},pdfauthor={Yubang Wang, Zezheng Huai, Yifei Zhang, Jinhao Wu, Yitong Chen, Haozhe Zhang, Sixian Li, Jingjing Gong, Zhenyang Lin, Ziyang Chen, Xipeng Qiu}}
\usepackage{url}
\title{Embodied Task Agents Can Perform Fine Manipulation through Tactile In-Context Learning}

\author{%
Yubang Wang\textsuperscript{1,2,3}, Zezheng Huai\textsuperscript{2,6}, Yifei Zhang\textsuperscript{2,5}, Jinhao Wu\textsuperscript{2,5},\\[2pt]
Yitong Chen\textsuperscript{1,2}, Haozhe Zhang\textsuperscript{2,4}, Sixian Li\textsuperscript{1,2}, Jingjing Gong\textsuperscript{2,$\dagger$},\\[2pt]
Zhenyang Lin\textsuperscript{3}, Ziyang Chen\textsuperscript{3}, Xipeng Qiu\textsuperscript{1,2,$\dagger$}}
\affiliation{%
\textsuperscript{1}Fudan University\quad
\textsuperscript{2}Shanghai Innovation Institute\quad
\textsuperscript{3}Tencent\\[2pt]
\textsuperscript{4}Zhejiang University\quad
\textsuperscript{5}Nanjing University\quad
\textsuperscript{6}Jilin University}
\contribution[\dagger]{Corresponding authors}

\checkdata[Project page]{\url{https://no-518.github.io/tactile-eta/}}

\abstract{Can embodied task agents use tactile experience to perform fine manipulation without task-specific model training? We investigate this capability through Tactile-ETA, a system that incorporates tactile experience into an embodied task agent through in-context learning. Successful expert demonstrations show how contact changes during manipulation. We pair each recorded robot movement with its camera views and bilateral tactile observations. We keep tactile observations at the start and end of each movement and add two neighboring time points when the tactile images change markedly in between. The agent combines these demonstrations with current vision, tactile observations, and proprioception to select actions through general-purpose robot tools, without updating model parameters. Across four simulated manipulation tasks, adding tactile observations to the same two expert demonstrations increases average success from 15.75\% to 24.00\%, with current sensing and execution interfaces held fixed. Further experiments show that performance depends on the sensory content and representation of demonstrations, while comparing agent models on Lift Bottle reveals substantial differences under the same system configuration. On the real-robot tasks, demonstrations with aligned visual and tactile observations increase the task success rate from 0\% to 60\%. These findings demonstrate the feasibility of using tactile experience in context for fine manipulation and suggest a path toward extending embodied task agents to additional sensory inputs.}

\begin{document}

\maketitle
\suppressfloats[t]

\begin{figure}[t]
\centering
\includegraphics[width=0.9\linewidth]{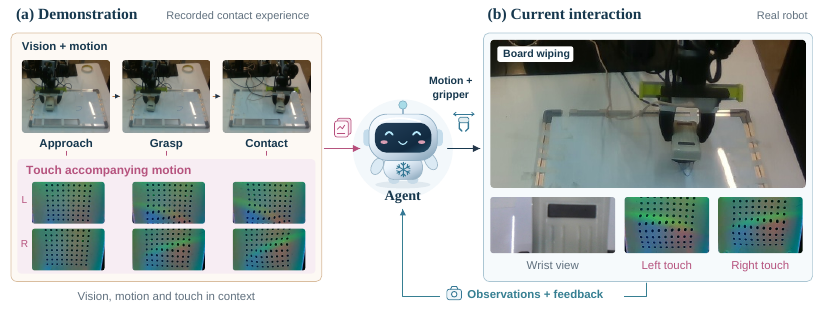}
\caption{Tactile-ETA performs fine manipulation by conditioning a frozen
embodied task agent on (a) action-aligned expert demonstrations that preserve
vision, motion, and tactile observations, together with (b) current observations and
execution feedback---without parameter updates.}
\label{fig:overview}
\end{figure}

\section{Introduction}

Embodied task agents such as ETA~\citep{chen2026etanewagenticparadigm},
Show-Harness~\citep{chen2026showharnessjustvlmagent}, and
RoboDawn~\citep{guo2026transferringintelligencevlmsrobotic}
demonstrate that general-purpose models can execute robot tasks through
observations, reasoning, and robot commands.
However, fine manipulation remains challenging.
Recent real-robot evaluation also shows a substantial performance
gap between pick-and-place tasks and fine insertion tasks
\citep{menon2026gpt6astra}.
Such manipulation requires adjusting actions to contact conditions,
which can be difficult to infer reliably from vision alone.
Tactile feedback provides complementary information for these adjustments
\citep{bi2025vlatouchenhancingvisionlanguageactionmodels}.
In-context learning further allows demonstrations to provide task guidance
at deployment without additional parameter updates
\citep{dipalo2024keypointactiontokensenable,
yin2025incontextlearningenablesrobot,cheng2026incontextrobotlearningvlm}.
This creates an opportunity to extend the sensory information these agents
can use: \emph{can embodied task agents similarly use tactile experience
in context to perform fine manipulation that requires contact-sensitive
adjustments?} We investigate this question through \textbf{Tactile-ETA}
(Figure~\ref{fig:overview}), which incorporates expert tactile experience
into an embodied task agent that acts through general-purpose robot tools,
without task-specific fine-tuning.

Current tactile
observations describe ongoing contact, whereas expert demonstrations can
provide a reference for how contact evolves during movement. For example,
during rotation or insertion, a demonstration can show how an expert
continues moving, adjusts its pose, or changes manipulation stages around
a contact transition. Visual--motion records capture what the expert did, but not how contact evolved during the demonstrated actions. We organize expert demonstrations so that the agent can see each movement
together with the visual and tactile changes that accompany it.

Tactile-ETA divides successful expert demonstrations into action-aligned
interaction segments, each containing robot motion, before-and-after
camera observations, and bilateral tactile sequences from the same interval.

Full tactile sequences can contain many observations within one movement,
while the endpoints alone can miss changes during contact. We therefore
retain tactile observations at motion boundaries and, when a change
threshold is met, add an adjacent pair spanning the selected change.
Complete demonstrations are provided in interaction order as references
for execution. The agent combines them with current visual and tactile observations, proprioception, and
execution feedback to select motion and
gripper commands. Model parameters remain unchanged: task guidance is
provided through context, while general-purpose tools and the underlying
controller execute the selected actions.

We evaluate the system on four UniVTAC simulation tasks
\citep{chen2026univtacunifiedsimulationplatform}, with 100 trials per task
and condition. Adding tactile observations to the same two visual--motion
demonstrations increases average success from 15.75\% to 24.00\%, with current
sensing and execution interfaces held fixed. Content ablations reveal
task-dependent contributions from demonstration vision and tactile
observations, while representation ablations show that retaining more
tactile frames does not automatically improve performance. A comparison of agent models
on Lift Bottle also reveals substantial differences under the
same system configuration. Execution statistics and cases illustrate
behavioral differences in contact-guided corrections and stage transitions.
On two real-robot tasks, a single multimodal demonstration enables three
successful trials out of five per task, compared with zero without demonstrations.
Together, these results demonstrate the feasibility of using tactile
experience in context for embodied task agents and suggest a path toward
extending them to additional sensory inputs.

Our contributions are threefold:
\begin{itemize}
    \item We introduce Tactile-ETA, a system that incorporates expert tactile
    experience into embodied task agents through context for fine manipulation,
    without task-specific parameter updates.
    \item We construct action-aligned visual--motion--tactile demonstrations
    that compactly retain motion boundaries and intermediate contact changes.
    \item We show that tactile demonstration experience can improve
    manipulation performance, with outcomes depending on context organization
    and the underlying agent model. Execution records illustrate associated
    differences in contact adjustment and stage transitions.
\end{itemize}

\Needspace{12\baselineskip}
\section{Related Work}
\label{sec:related}
\paragraph{Embodied Task Agents.}
Language and vision-language models connect task reasoning to robot execution through skill selection, environmental feedback, and program generation~\citep{ahn2022icanisay,huang2022innermonologueembodiedreasoning,liang2023codepolicieslanguagemodel}. Visual grounding translates model outputs into spatial targets, objectives, or constraints for robot execution~\citep{huang2023voxposercomposable3dvalue,huang2024rekepspatiotemporalreasoningrelational,liu2024mokaopenworldroboticmanipulation,nasiriany2024pivotiterativevisualprompting,huang2024copageneralroboticmanipulation}. Recent work studies general-purpose models operating through robot interfaces~\citep{fu2026capxframeworkbenchmarkingimproving,chen2026etanewagenticparadigm,chen2026showharnessjustvlmagent,wang2025wonderfulteamzeroshotphysical,su2026astra}. Related systems improve robot programs, skills, or policies through agent-driven iteration~\citep{chen2026gapgraphaspolicymultiagentselflearning,lu2026aspireagenticskillsdiscovery,xiao2026enpireagenticrobotpolicy}. We build on OpenETA's tool-based execution~\citep{chen2026etanewagenticparadigm}, in which the agent selects motion and gripper commands and receives observations and feedback after execution.

\paragraph{In-Context Robot Learning.}
Demonstration-guided robot learning~\citep{ARGALL2009469} includes methods that use examples at deployment without further parameter updates~\citep{jiang2023vimageneralrobotmanipulation,dipalo2024dinobotrobotmanipulationretrieval,Heppert_2024}. Some methods train policies to predict actions conditioned on demonstration trajectories~\citep{fu2024incontextimitationlearningnexttoken,vosylius2025instantpolicyincontextimitation}. For frozen general-purpose models, demonstration context can encode visual keypoints and end-effector trajectories as tokens~\citep{dipalo2024keypointactiontokensenable}, or initial object poses and keyframe actions as text~\citep{yin2025incontextlearningenablesrobot}. More closely related to our setting, visual demonstrations and action references are combined with online observations and execution feedback for closed-loop manipulation~\citep{cheng2026incontextrobotlearningvlm}. Interface-aligned demonstrations further illustrate motion-command effects and task strategies~\citep{guo2026transferringintelligencevlmsrobotic}.

\paragraph{Tactile Sensing for Robotic Manipulation.}
Tactile learning supports physical perception and action generation~\citep{guzey2023touchlearningtactiledexterity}. Existing work learns transferable tactile representations~\citep{pmlr-v270-higuera25a}, connects tactile observations to language-based reasoning~\citep{yu2024octopiobjectpropertyreasoning,yu2025demonstratingoctopi15visualtactilelanguagemodel}, and learns sequential visuotactile representations for manipulation~\citep{xu2026vtmusemultimodalunifiedsequential}. Policy learning incorporates tactile information through model training~\citep{hao2025tlatactilelanguageactionmodelcontactrich,huang2026tafvlatactileforcealignmentvisionlanguageaction}, with generalist tactile policies extending across sensors~\citep{yuan2026ftp1generalistfoundationtactile}. Semantic tactile feedback can also support planning alongside a learned controller for action refinement~\citep{bi2025vlatouchenhancingvisionlanguageactionmodels}. Particularly relevant, visual, tactile, and force--torque demonstrations have been used to infer skill sequences and refine their transition conditions, producing a reference plan for subsequent execution and replanning~\citep{chen2025lemmoplanllmenhancedlearningmultimodal}. Our approach retains expert motions and aligned visual and bilateral tactile observations as demonstration context for successive tool-based decisions. We study how this contact experience can guide an embodied task agent, including controlled comparisons that vary demonstrated tactile information while holding current sensing and execution interfaces fixed.

\section{Tactile-ETA}
\label{sec:method}

\begin{figure}[t]
\centering
\includegraphics[width=0.95\linewidth]{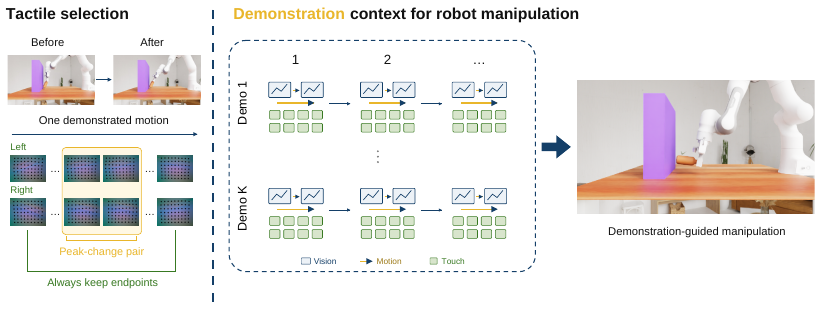}
\caption{Tactile context construction. (Left) Keep tactile endpoints and,
when the threshold is met, the peak-change pair. (Right) Combine tactile observations,
vision, and recorded robot movements into ordered demonstrations for robot manipulation.}
\label{fig:method}
\end{figure}

\subsection{Tactile-ETA Overview}
\label{sec:formulation}

Tactile-ETA incorporates tactile experience into an embodied task agent
through context, without task-specific parameter updates
(Figure~\ref{fig:overview}). The agent receives a task instruction,
expert demonstrations, and current visual, tactile, and proprioceptive
observations. Expert demonstrations provide the agent with a reference for how the task is successfully executed, including both the expert actions and the contact changes that accompany them, while current observations and execution
feedback describe the ongoing interaction. The agent uses both to select
calls to general-purpose robot tools; an underlying controller executes
the requested movements.

Let $\mathcal C_K$ denote the ordered context constructed from $K$
expert episodes, $\ell$ the task instruction, and $\mathcal I$ the tool
descriptions and static task requirements. Before tool call $t$, the
current-episode history $H_t$ contains the observations received so far,
previous tool calls, and execution feedback. A model with fixed parameters
$\theta$ selects a tool name and arguments $u_t$ according to
\begin{equation}
    u_t \sim \pi_\theta\!\left(
        \cdot \mid \ell,\mathcal I,\mathcal C_K,H_t
    \right).
    \label{eq:tool-call}
\end{equation}
The demonstration context is provided once at the start of the episode
and remains fixed. After each tool call, returned observations and feedback
extend $H_t$ for the next decision, without changing $\theta$. The following
sections describe how we construct the tactile demonstration context and
make current tactile feedback available during execution.

\subsection{Tactile Experience in Context}
\label{sec:demonstration-context}
\label{sec:tactile-selection}

We construct demonstration context that links expert movements to the
visual and tactile observations accompanying them (Figure~\ref{fig:method}).
Each successful expert episode is divided at its recorded action boundaries.
Within a segment, before-and-after head- and wrist-camera views describe
the scene, while synchronized bilateral tactile observations capture the
contact process. The motion record contains measured end-effector
translation, relative rotation, and finger-position changes computed from
the recorded robot states. These motions and the visual and tactile
observations refer to the same action interval.
Record fields, units, and coordinate conventions are provided in
Appendix~\ref{app:demonstration-construction}.

\paragraph{Compact tactile sequences.}
We retain tactile observations at both action boundaries.
For each fingertip, we select small patches in one tactile image
and find their most similar matches at nearby locations in the next image.
The distance between each patch's original and matched locations gives
its displacement. We sort the displacement magnitudes of reliable matches
from smallest to largest and use the value approximately three quarters
of the way through this list, the 75th percentile $d_{75}$, to measure
the change between the two images.
We express $d_{75}$ in original-image pixels and divide it by the
selection threshold of 2 pixels to obtain a normalized change score.
When too few patches can be matched reliably, we instead use an
image-difference score normalized by its own selection threshold
(Appendix~\ref{app:demonstration-construction}).

For a segment with $N\geq2$ synchronized samples per fingertip, let $c_n$
be the larger normalized score across the two fingertips between
samples $n-1$ and $n$.
In addition to the endpoints, we retain at most one adjacent pair
per segment: the pair with the largest $c_n$, provided that this
score is at least one.
Let
\[
    n^\star = \underset{1\leq n<N}{\operatorname{arg\,max}}\;c_n.
\]
We retain the time indices
\begin{equation}
    \mathcal S =
    \begin{cases}
        \{0,\,n^\star-1,\,n^\star,\,N-1\}, & c_{n^\star}\geq1,\\[4pt]
        \{0,\,N-1\}, & c_{n^\star}<1.
    \end{cases}
    \label{eq:tactile-frame-selection}
\end{equation}
The endpoints associate tactile appearances with the before-and-after
visual states; the additional pair preserves an intermediate transition
within the same movement.
Both fingertips use the same indices, which are sorted and deduplicated,
leaving at most four observations per fingertip. A single-sample segment
retains its sole observation. The selected images are presented as separate
left and right chronological strips with recorded step indices.
Selection parameters and an example are given in
Appendix~\ref{app:demonstration-construction}.

\paragraph{Context assembly.}
Each segment pairs these tactile strips with its visual observations and
motion record. Segments retain their order within each complete expert
episode, together with the recorded episode outcome. The resulting $K$
episodes form $\mathcal C_K$, providing a reference for what the expert did
and how contact evolved during those movements.

\subsection{Execution with Current Tactile Feedback}
\label{sec:closed-loop}

At the beginning of an episode, the embodied task agent receives the
demonstration context and selects actions using the task instructions,
current visual and tactile observations, and robot state.
It interacts through OpenETA's general-purpose robot
tools~\citep{chen2026etanewagenticparadigm}.
The \texttt{move\_to} tool specifies an end-effector position and
orientation target, together with an optional gripper command;
the underlying controller executes the movement.
The \texttt{observe} tool returns current camera images, bilateral
tactile observations, end-effector poses, and joint positions.
Demonstrations provide a reference for choosing actions in the current scene.

After each motion segment, the system returns the latest camera images
and robot state, along with feedback on whether the target was reached
and the deviation from the requested motion.
Tactile observations recorded during the movement are selected using
the rule in Section~\ref{sec:tactile-selection} and presented as
separate left and right image strips.
The tool call and its returned observations and feedback are appended
to the interaction history for subsequent action selection.
The agent can therefore refer to both the contact changes in expert
demonstrations and the latest feedback from its own actions.
Model parameters remain fixed throughout execution.
Operator prompts, tool descriptions, and the motion argument schema are provided in
Appendix~\ref{app:tools-and-prompts}.

\section{Experiments}
\label{sec:experiments}

We evaluate whether embodied task agents can perform fine manipulation
through tactile in-context learning. We examine how demonstrations guide
action selection and how performance depends on demonstration content,
tactile sequence representation, and the agent model.

\paragraph{Experimental setup.}
We evaluate on four tasks from
UniVTAC~\citep{chen2026univtacunifiedsimulationplatform}:
Insert Tube, Lift Can, Lift Bottle, and Pull Out Key.
Corresponding comparisons use the same 100 evaluation initial conditions per
task, with success and automatic termination determined by the native
environment. Unless otherwise specified, we use a frozen GPT-6 agent
with \texttt{low} reasoning effort and provide two successful expert
episodes from the official dataset when demonstrations are used.
Demonstration identities and ordering, robot tools, controller, task
instructions, and execution budgets are held fixed within each comparison.
Full settings are provided in Appendix~\ref{app:details}.

\subsection{Fine Manipulation with Tactile Experience}
\label{sec:simulation}

\paragraph{Compared configurations.}
We compare four input configurations of the same embodied task agent.
All receive current vision, proprioception, and execution feedback.
\emph{No demos (vision)} provides neither demonstrations nor tactile
observations. \emph{No demos (vision + tactile)} adds current tactile
observations. \emph{Visual--motion demos} provides current tactile
observations and two expert demonstrations containing visual observations
and motion records. \emph{Tactile-ETA} further includes the tactile
observations from those demonstrations.
The latter two configurations use the same expert episodes and current
sensory inputs, differing only in whether the demonstrations contain
tactile observations.

\suppressfloats[t]
\begin{table}[!t]
    \centering
    \caption{
        Success rates (\%) on four UniVTAC tasks, with 100 trials
        per task and condition.
        Bold denotes the best result in each row.
    }
    \label{tab:main}
    \small
    \renewcommand{\arraystretch}{1.15}
    \begin{tabular}{@{}lcccc@{}}
        \toprule
        Task
        & \shortstack{No demos\\(vision)}
        & \shortstack{No demos\\(vision + tactile)}
        & \shortstack{Visual--motion\\demos}
        & \shortstack{Tactile-ETA\\(ours)} \\
        \midrule
        Insert Tube
        & 3  & 15 & 15 & \textbf{32} \\
        Lift Can
        & 0  & 1  & \textbf{3} & 2 \\
        Lift Bottle
        & 7  & 3  & 27 & \textbf{36} \\
        Pull Out Key
        & 2  & 6  & 18 & \textbf{26} \\
        \midrule
        Average success (\%)
        & 3.00
        & 6.25
        & 15.75
        & \textbf{24.00} \\
        \bottomrule
    \end{tabular}
\end{table}
\Needspace{7\baselineskip}
\paragraph{Results.}
Adding tactile observations to visual--motion demonstrations increases
average success across the four tasks from 15.75\% to 24.00\%, a gain
of 8.25 percentage points (Table~\ref{tab:main}). With current sensory
inputs held fixed, this comparison shows that expert tactile experience
provides additional guidance for fine manipulation.
For reference, average success without demonstrations is 3.00\% with
current vision alone and 6.25\% with current vision and tactile sensing.
On Insert Tube, visual--motion demonstrations provide no success-rate gain
over the agent with current vision and tactile sensing; adding demonstrated
tactile observations improves performance. On Lift Bottle and Pull Out Key,
visual--motion demonstrations already improve success, and including
demonstrated tactile observations produces a further gain. These results
show that visual--motion guidance and demonstrated tactile experience
contribute differently across tasks. Lift Can remains difficult under all
configurations.

\subsection{How Tactile Demonstrations Guide Manipulation}
\label{sec:behavioral-analysis}

Demonstrations help the agent establish an initial manipulation strategy.
On Lift Bottle, the agent with current vision and tactile sensing but no
demonstrations begins by closing the gripper in place in only 1/100
episodes, compared with 100/100 under visual--motion guidance.
Tactile-ETA adopts the same initial strategy.
Expert visual and motion records therefore already provide a concrete
starting action for this task.

\begin{figure}[t]
\centering
\includegraphics[width=\linewidth]{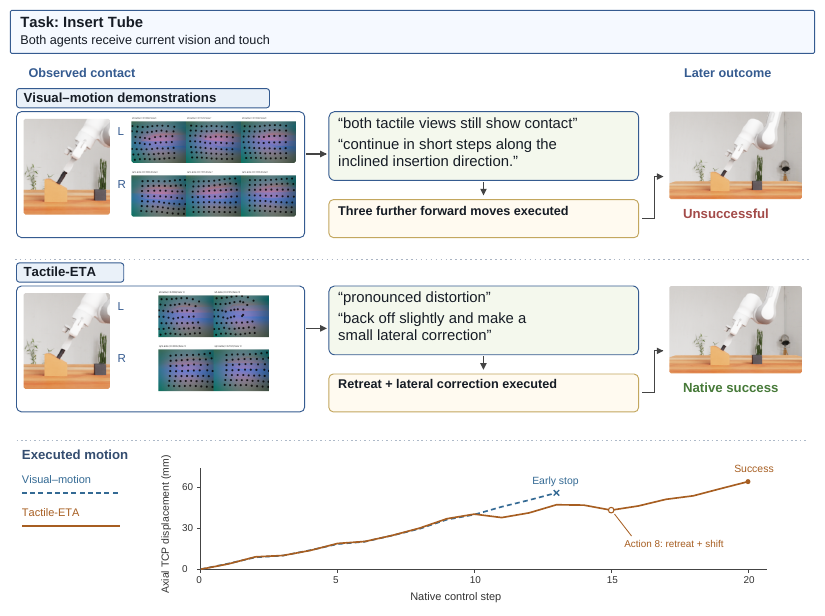}
\caption{Two Insert Tube rollouts with different demonstration contexts.
Visual--motion demonstrations lead to continued advancement and failure;
tactile-augmented demonstrations guide retreat and lateral correction,
followed by success. Curves show measured TCP displacement along each
rollout's initial approach direction.}
\label{fig:rollouts}
\end{figure}

Tactile experience in demonstrations also influences adjustments after
contact. On Insert Tube, 48/100 visual--motion episodes and 13/100
Tactile-ETA episodes end in unsuccessful native early termination with
at least two executed segments whose net TCP displacement along the
initial approach axis exceeds 1\,mm, and no segment with a net displacement
below $-1$\,mm along that axis (Appendix~\ref{app:tube-decisions}).
Figure~\ref{fig:rollouts} shows two examples: the former continues
requesting forward motion after observing contact and ultimately fails;
the latter reports pronounced tactile distortion, executes a retreat and
lateral correction, and subsequently completes insertion.
Both agents receive current tactile feedback.
Together with the success-rate comparison, these behavioral differences
support the use of demonstrated tactile experience to guide action
adjustments in response to contact.

\subsection{What Makes Tactile Context Effective?}
\label{sec:ablations}

We examine demonstration content and tactile sequence representation.
The content and representation comparisons use the same two expert episodes and 100 evaluation initial
conditions per task, with current vision, tactile observations,
proprioception, and execution feedback held fixed.
Visual--motion and Tactile-ETA/Selected results are reused from
Table~\ref{tab:main}.

\Needspace{14\baselineskip}
\begin{wraptable}{r}{0.51\linewidth}
    \centering
    \caption{Demonstration content ablation. Success rates (\%); bold marks the best per task.}
    \label{tab:ablation-designs}
    \small
    \setlength{\tabcolsep}{3pt}
    \renewcommand{\arraystretch}{1.15}
    \begin{tabular*}{\linewidth}{@{}l@{\extracolsep{\fill}}cc@{}}
        \toprule
        Context & \shortstack{Lift\\Bottle} & \shortstack{Pull Out\\Key} \\
        \midrule
        Motion records & 24 & 10 \\
        Visual--motion & 27 & 18 \\
        Tactile--motion & 26 & \textbf{26} \\
        Tactile-ETA (ours) & \textbf{36} & \textbf{26} \\
        \bottomrule
    \end{tabular*}
\end{wraptable}

\paragraph{Demonstration content.}
We retain the motion records and compare contexts with no sensory
observations, visual observations, tactile observations, or both.
Table~\ref{tab:ablation-designs} shows that the contribution of sensory
information varies across tasks. On Pull Out Key, tactile--motion
demonstrations achieve the same observed success rate as the complete
representation. On Lift Bottle, the complete representation performs best,
indicating that visual information remains a useful reference for this task.

Figure~\ref{fig:app-demonstration-content} in the appendix illustrates
different stage transitions under two demonstration configurations.
Both agents follow a rotate--reverse--lift procedure.
The agent given motion records alone reverses after its third forward
rotation and fails during extraction. The tactile--motion agent performs
another rotation after reporting no clear stopping cue, then reverses
and pulls after a pronounced tactile change, completing the task.
These examples illustrate the distinction between the action sequence
and the timing of stage transitions, and show how the agent adjusts
execution in response to contact changes.

\Needspace{14\baselineskip}
\begin{wraptable}{r}{0.46\linewidth}
    \centering
    \caption{Tactile representation ablation on Lift Bottle. Success rates (\%).}
    \label{tab:tactile-representation}
    \small
    \setlength{\tabcolsep}{3pt}
    \renewcommand{\arraystretch}{1.15}
    \begin{tabular*}{\linewidth}{@{}l@{\extracolsep{\fill}}r@{}}
        \toprule
        Representation & Success (\%) \\
        \midrule
        No demo tactile & 27 \\
        Endpoints & 27 \\
        All frames & 2 \\
        Selected (ours) & \textbf{36} \\
        \bottomrule
    \end{tabular*}
\end{wraptable}

\paragraph{Tactile sequence representation.}
On Lift Bottle, we compare \emph{Endpoints}, retaining only the first
and last tactile frames of each demonstrated action; \emph{All frames},
retaining the full recorded sequence; and \emph{Selected}, using the
selection rule in Section~\ref{sec:tactile-selection}.
The expert episodes, action segmentation, visual observations, and
motion records are unchanged.
As shown in Table~\ref{tab:tactile-representation}, Selected achieves
36\% success, compared with 27\% for Endpoints and 2\% for All frames.
Endpoints matches the success rate of demonstrations without tactile
observations. These results show that performance depends on how
tactile experience is represented in context.

\begin{figure}[!t]
\centering
\includegraphics[width=\linewidth]{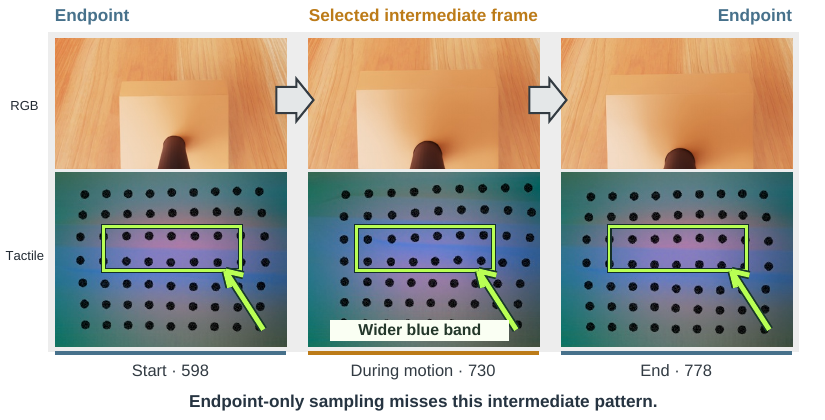}
\caption{\textbf{Intermediate tactile information missed by endpoint-only sampling.}
Columns show the start, a selected intermediate frame, and the end of an
Insert Tube demonstration segment, with synchronized RGB observations above
and right-fingertip tactile images below. Although the endpoint tactile
patterns look similar, the intermediate frame exhibits a wider blue band
in the highlighted region. Endpoint-only sampling omits this intermediate
pattern, whereas Selected retains it in the demonstration context.
Green boxes and arrows identify the same image region for comparison. One of the two selected
intermediate frames is shown.}
\label{fig:app-tactile-representation}
\end{figure}

Figure~\ref{fig:app-tactile-representation} illustrates the intermediate
tactile information retained by Selected in an Insert Tube demonstration.
Despite similar endpoint patterns, the selected intermediate frame shows
a wider blue band, revealing a change that endpoint-only sampling omits.

\WFclear

With one, two, and four complete demonstrations, mean success rates are
22.50\%, 24.00\%, and 19.00\%, respectively
(Appendix~\ref{app:additional-ablations}). Increasing from two to four
raises Pull Out Key success from 26\% to 48\% but reduces Lift Bottle
success from 36\% to 2\%.

\Needspace{20\baselineskip}
\subsection{Dependence on the Agent Model}
\label{sec:agent-backbone}

\begin{wrapfigure}[16]{r}{0.49\linewidth}
\centering
\includegraphics[width=\linewidth]{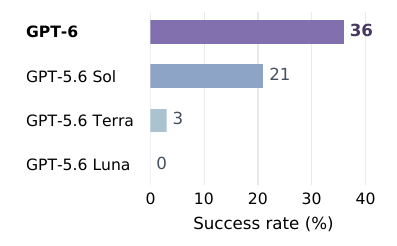}
\caption{Agent model comparison on Lift Bottle: 100 trials per model with the same two demonstrations. GPT-6 results are reused from Table~\ref{tab:main}.}
\label{fig:agent-backbone}
\end{wrapfigure}
We compare four agent models on Lift Bottle using the same two
tactile-augmented demonstrations and 100 evaluation initial conditions.
All models use the \texttt{low} reasoning-effort setting, with the
demonstration representation, task instructions, observation modalities,
robot tools, controller, and execution budgets held fixed.

Figure~\ref{fig:agent-backbone} shows substantial differences in
manipulation performance despite identical demonstrated experience and
execution interfaces. These results indicate that tactile context design
should be considered together with the agent's ability to use observations,
demonstrations, and feedback for decision-making.

\subsection{Real-World Experiments}
\label{sec:realworld}
\WFclear

\begin{figure}[H]
\centering
\includegraphics[width=\linewidth]{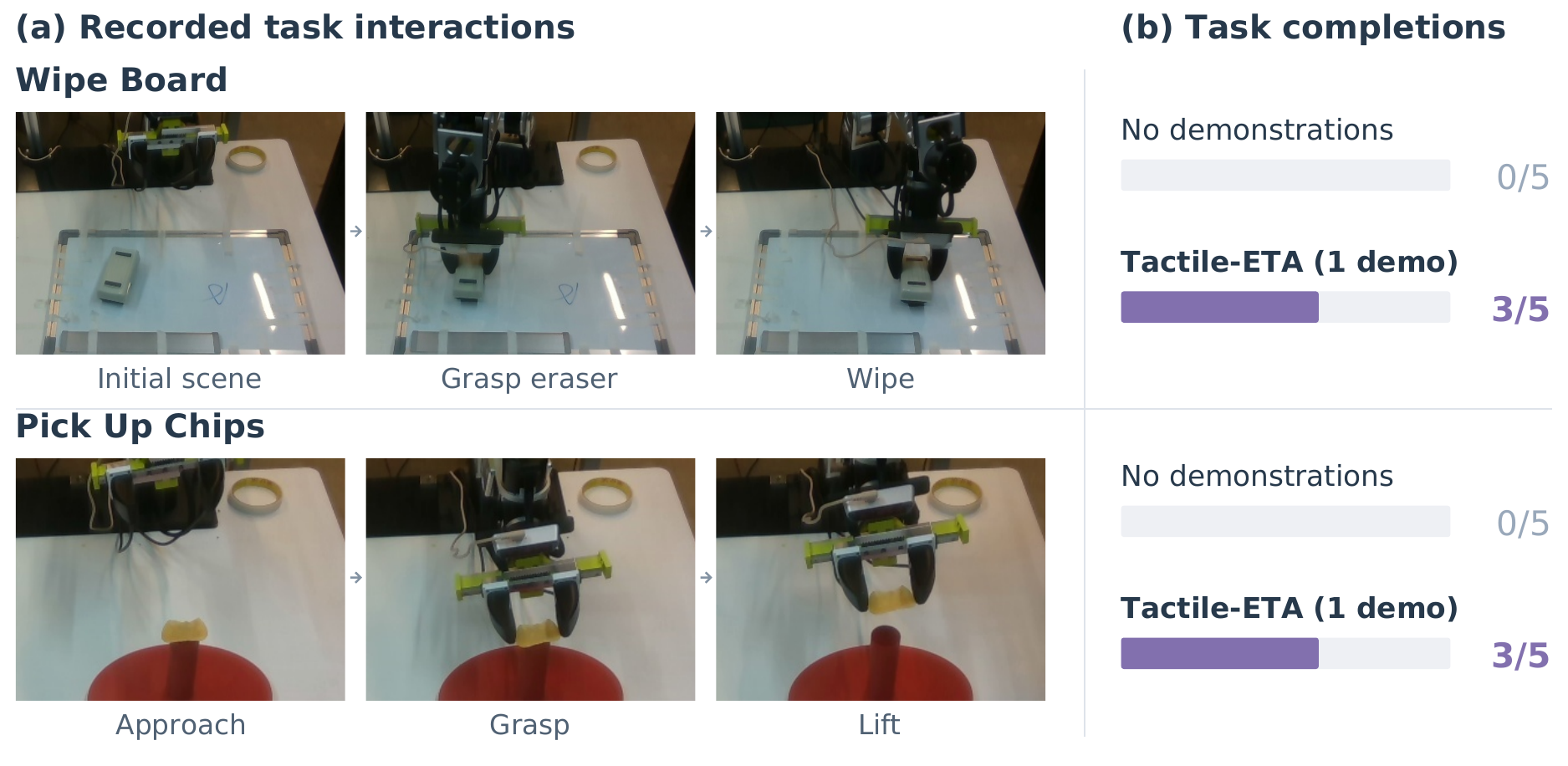}
\caption{Real-world tasks and results. (a) Recorded interactions illustrate wiping and chip pickup. (b) Task completions over five trials per configuration. Both receive current vision and tactile observations; Tactile-ETA additionally receives one multimodal human demonstration.}
\label{fig:realworld}
\end{figure}

We further evaluate Tactile-ETA on a real robot with two tasks:
\emph{Wipe Board} and \emph{Pick Up Chips}.
The platform uses a reBot Arm B601-RS equipped with a vitai GF225 tactile
gripper.
Visual observations come from an Orbbec wrist-mounted camera and a
RealSense D435i third-person camera.
The former requires wiping a whiteboard to remove the target writing;
the latter requires lifting a potato chip without breaking or dropping it.
We compare execution without demonstrations against execution with one
visual--motion--tactile demonstration. Both configurations receive current
vision, tactile observations, and robot state. The demonstration is provided
through context, with model parameters held fixed.
Figure~\ref{fig:realworld}(a) illustrates the two tasks.
A video of the Wipe Board experiment is available on the \href{https://no-518.github.io/tactile-eta/}{project website}.
Detailed settings and success criteria are provided in
Appendix~\ref{app:realworld-details}.

As shown in Figure~\ref{fig:realworld}(b), the agent without demonstrations
completes neither task in any of the five trials, while Tactile-ETA completes
three of five trials on each task. These results extend our findings on
demonstration experience to real robots: with current tactile sensing
available, a single multimodal demonstration provides useful guidance for
manipulation. Together with the controlled comparisons of demonstrated
tactile information in simulation, the real-world results further demonstrate
the feasibility of incorporating tactile experience into embodied task agents
through context.

\section{Discussion}
\label{sec:discussion}

\paragraph{Limitations and future directions.}
\label{sec:limitations}
Tactile-ETA explores how incorporating touch as an additional input modality through in-context experience can extend embodied task agents' capabilities for fine manipulation. We hope this work encourages broader exploration of how new sensory modalities can expand agents' manipulation capabilities. Our evaluation is limited in model coverage by API availability and inference cost; broader community evaluation could help characterize how different foundation models use tactile experience. In addition, the current system primarily uses tactile feedback to guide decisions between motion segments. Combining decisions informed by tactile experience with fast tactile feedback control during action execution is a promising direction for improving continuous contact-rich interaction.

\paragraph{Conclusions.}
\label{sec:conclusion}
We introduced Tactile-ETA, which enables embodied task agents to use tactile experience in context for fine manipulation without updating model parameters. The system organizes expert demonstrations into compact, action-aligned visual, motion, and tactile context and combines this experience with current sensing and execution feedback. Across four simulated tasks, adding tactile observations to visual–motion demonstrations improves average success while retaining the same current sensory inputs. Further experiments show that performance depends on demonstration content, tactile representation, and the agent model. Real-robot experiments demonstrate successful wiping and fragile-object pickup with a single multimodal demonstration. These findings establish the feasibility of tactile in-context learning for fine manipulation and suggest a path toward extending the sensory capabilities of embodied task agents through experience provided in context.
\label{maintext:end}

\bibliography{iclr2027_conference}
\bibliographystyle{plainnat}

\appendix
\clearpage
\section{Experimental Setup}
\label{app:details}

We detail the simulation protocol and controlled inputs here.
Appendix~\ref{app:demonstration-construction} describes demonstration
construction; Appendix~\ref{app:tools-and-prompts} provides the operator
prompts and tool schemas.

\subsection{Simulation setup}
\label{app:simulation-setup}

We evaluate the main input configurations on four UniVTAC tasks:
Insert Tube, Lift Can, Lift Bottle, and Pull Out Key.
Each configuration is evaluated on the same 100 initial conditions per
task, using separate rollouts.
The agent begins operating after the environment's native reset
and preparatory motion. From this point onward, it selects the
motion and gripper commands through the robot tools.
Success rates are computed using the native task criteria and
reported both per task and averaged across tasks.

The main comparisons use a frozen GPT-6 agent with the
\texttt{low} reasoning-effort setting. Demonstration-based
conditions receive the same two successful official expert
episodes, 0 and 1, in that order. One demonstration denotes a
complete expert episode rather than a single action segment.
The demonstration-content ablation covers Lift Bottle and
Pull Out Key. The tactile-representation comparison in
Table~\ref{tab:tactile-representation} concerns Lift Bottle, and
the model comparison is reported in Section~\ref{sec:agent-backbone}.

\subsection{Input configurations}
\label{app:input-configurations}

Table~\ref{tab:appendix-inputs} summarizes the information available
in the main comparisons and demonstration-content ablations.
All configurations retain current camera observations,
proprioception, and ordinary execution feedback. Removing images
from the demonstrations does not remove the corresponding current
observations. Only the no-demonstration vision baseline omits
current tactile observations.

\begin{table}[H]
    \centering
    \caption{Input configurations. Numerical demonstration records
        contain recorded robot movements, robot states, timing, and episode
        outcomes. All settings retain current camera observations,
        proprioception, and ordinary execution feedback. Motion records
        and Tactile--motion are additional content ablations evaluated
        on Lift Bottle and Pull Out Key.}
    \label{tab:appendix-inputs}
    \small
    \renewcommand{\arraystretch}{1.12}
    \begin{tabular*}{\linewidth}{@{\extracolsep{\fill}}lcccc@{}}
        \toprule
        Setting & \shortstack{Current\\tactile}
        & \shortstack{Demonstration\\records}
        & \shortstack{Demonstration\\vision}
        & \shortstack{Demonstration\\tactile} \\
        \midrule
        No demos (vision)         & --  & --  & --  & --  \\
        No demos (vision + tactile) & Yes & --  & --  & --  \\
        Motion records           & Yes & Yes & --  & --  \\
        Visual--motion           & Yes & Yes & Yes & --  \\
        Tactile--motion           & Yes & Yes & --  & Yes \\
        Tactile-ETA               & Yes & Yes & Yes & Yes \\
        \bottomrule
    \end{tabular*}
\end{table}

For the four demonstration-content settings, the expert episodes,
segment boundaries, recorded robot movements, robot states, timing, and
outcomes remain the same. Only the images in the demonstrations are varied.
The tactile--motion condition retains the same bilateral tactile
sequences as the complete representation, while omitting the
camera images and their references from the demonstrations.

\subsection{Execution settings}
The agent issues motion targets through \texttt{move\_to} and obtains
observations through \texttt{observe}. Camera views, robot state, enabled
tactile history, and target-tracking feedback are returned after execution.
Table~\ref{tab:execution-settings} lists the controller and episode budgets;
exact interface conventions are given in Appendix~\ref{app:tools-and-prompts}.

\begin{table}[H]
\centering\small
\caption{Execution settings in the retained prompt and configuration records.}
\label{tab:execution-settings}
\begin{tabular}{ll}
\toprule
Setting & Value \\
\midrule
Physics / control frequency & 120 / 60 Hz (decimation 2) \\
Maximum control steps per motion & 80 \\
Admitted non-preview motions & 30 per episode \\
MCP calls & 100 per episode \\
Total operator wall time & 3600 s \\
Post-terminal observation/completion grace & Up to 300 s within total budget \\
Insert Tube / Lift Can control-step limit & 300 / 300 \\
Lift Bottle / Pull Out Key control-step limit & 500 / 300 \\
Position / orientation arrival tolerance & 0.003 m / 0.05236 rad \\
Maximum joint change per control step & 0.025 rad \\
Gripper settle time / tolerance & 0.1 s / 0.0002 m \\
Gripper target tolerance / maximum wait & 0.001 m / 40 control steps \\
\bottomrule
\end{tabular}
\end{table}

\subsection{Native evaluation and termination}
\label{app:evaluation}

Native success checks, early-failure conditions, and episode
budgets determine when physical interaction ends. These checks
run in the background. The agent is not given current task
scores or the specific native termination reason, and no
\texttt{check\_task} tool is available. Automatic termination
is reported with the neutral notification
\texttt{episode\_ended}. The \texttt{finish\_episode} tool is
used to end voluntarily or acknowledge termination.
Historical expert outcomes and static task requirements remain
available; they are distinct from evaluation of the current
rollout. Success rates use the environment's recorded labels,
not the agent's final account of its performance.

We retain the benchmark's task definitions. In particular,
Lift Can is evaluated using its native pose predicate, not a
separate criterion for stable suspended lifting.

\section{Demonstration Construction}
\label{app:demonstration-construction}

Each recorded action segment pairs robot movement and boundary robot states
with before-and-after camera views and selected bilateral tactile observations, following
Section~\ref{sec:demonstration-context}. Complete expert episodes retain
segment order and recorded outcomes. The same two episodes are supplied
once at rollout start. We first show a complete recorded action segment, then specify tactile
selection and the formats used to present the resulting observations.

\subsection{An action-segment example}
\label{app:segment-example}

We show segment 5 of Lift Bottle expert episode 0, spanning recorded steps
392--464, with boundary positions, measured movement, camera observations,
and bilateral tactile sequences. This is one segment of a demonstration,
not a complete episode or the entire two-demonstration tool response.
Numerical values below are rounded for readability.

\begin{appendixprompt}{Lift Bottle: episode 0, segment 5}
\textbf{Boundary states.} Steps 392 $\rightarrow$ 464; hand position (m):
$[0.55427,-0.04652,0.21738] \rightarrow [0.57888,-0.04511,0.24504]$.
\par\smallskip
\textbf{Measured movement.} Translation: $[24.61,1.41,27.65]$ mm;
rotation vector: $[-4.89,2.40,2.88]\times10^{-6}$ rad;
finger changes: $[1.68,-1.68]\times10^{-8}$ m; nominal duration: 0.60 s.
\par\smallskip
\textbf{Tactile times.} $[392,428,430,464]$, shared by both fingertips.
\par\smallskip
\textbf{Feedback.} ``Observed state transition; original execution
feedback is not separately recorded.''
\par\medskip
\begin{minipage}[t]{0.48\linewidth}
{\footnotesize\ttfamily example 0 segment 5\par head before/after}\par
\includegraphics[width=\linewidth]{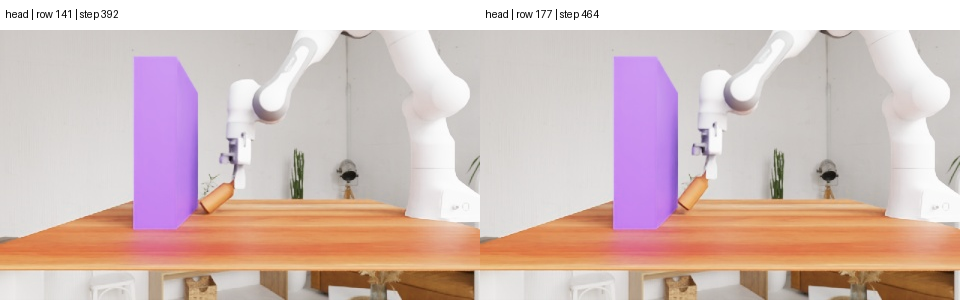}
\end{minipage}\hfill
\begin{minipage}[t]{0.48\linewidth}
{\footnotesize\ttfamily example 0 segment 5\par wrist before/after}\par
\includegraphics[width=\linewidth]{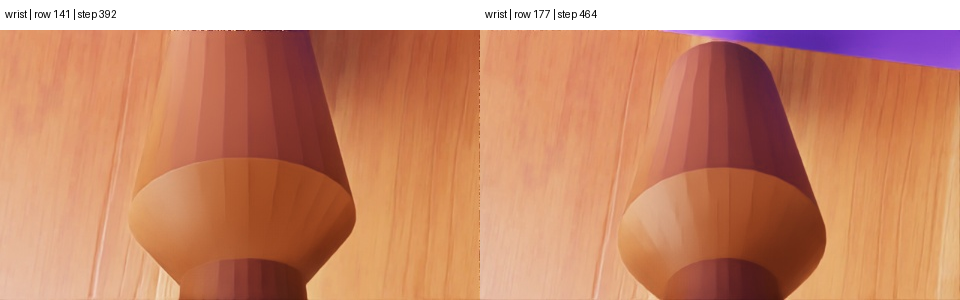}
\end{minipage}\par\smallskip
\begin{minipage}[t]{0.48\linewidth}
{\footnotesize\ttfamily example 0 segment 5\par left\_tactile history}\par
\includegraphics[width=\linewidth]{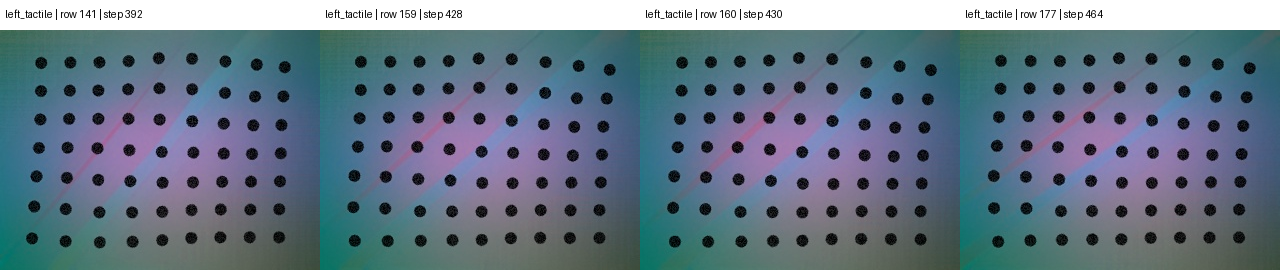}
\end{minipage}\hfill
\begin{minipage}[t]{0.48\linewidth}
{\footnotesize\ttfamily example 0 segment 5\par right\_tactile history}\par
\includegraphics[width=\linewidth]{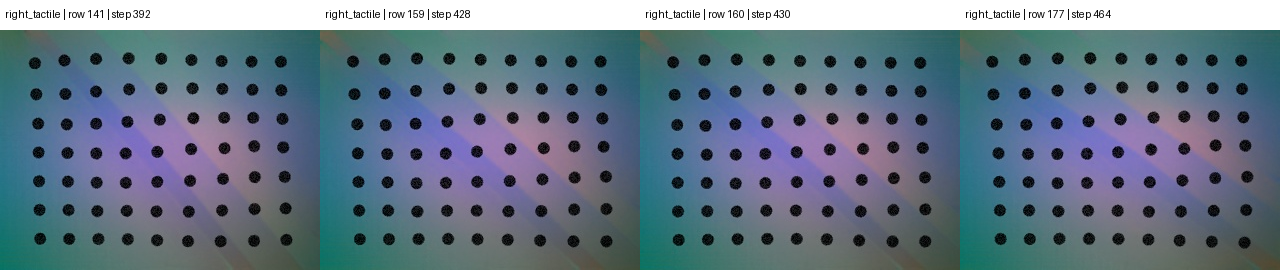}
\end{minipage}\par\smallskip
\end{appendixprompt}

Historical poses use robot-base-relative \texttt{panda\_hand} positions in
metres and quaternions in \texttt{wxyz} order. Joint states contain seven
arm angles in radians and two finger positions in metres. Nominal duration
is the recorded step difference divided by 120, rather than a measured
wall-clock interval. These measured movements are not recorded online
TCP commands.

The headings reproduce the image labels referenced by the segment's
\texttt{vision\_labels} and its associated \texttt{historical\_touch} record.
The tool interface supplies these labels in text alongside the image
blocks; local image paths are packaging locators and are omitted from the
transmitted descriptors. The strips above are the prepared input images,
including their recorded-step labels. They preserve the two camera
boundaries and the common four-time tactile sequence without adding
force or slip annotations.

\subsection{Tactile selection parameters}
Images are converted to grayscale and resized while preserving their aspect
ratio. We track a $3\times4$ grid of patches inside the region of interest
(ROI). Each patch is matched by an integer displacement search minimizing
mean squared error (MSE). Low-texture patches and ambiguous or high-error
matches are rejected. Tracking quality is the fraction of the 12 patches
that survive. The motion score is the 75th percentile of their displacement
magnitudes, rescaled to the original image resolution. When tracking quality
is below 0.35, we instead use the mean absolute grayscale difference inside
the ROI, divided by 255. Table~\ref{tab:touch-parameters} gives the retained
selection parameters for the expert demonstrations and online tactile history.

\begin{table}[H]
\centering\small
\caption{Tactile selection parameters. Spatial tracking settings use resized
pixels; the motion threshold uses original-image pixels.}
\label{tab:touch-parameters}
\begin{tabular}{ll}
\toprule
Parameter & Value \\
\midrule
Tracking width & 160 pixels \\
ROI $(x_0,y_0,x_1,y_1)$ & $(0.08,0.08,0.92,0.92)$ of image size \\
Tracking grid & 3 rows $\times$ 4 columns \\
Patch radius / patch size & 6 / $13\times13$ pixels \\
Search radius & 4 pixels ($9\times9$ candidates) \\
Minimum patch standard deviation & 8.0 grayscale levels \\
Maximum match MSE & 900.0 \\
Minimum second-best minus best MSE & 1.0 \\
Minimum tracking quality & 0.35 \\
Motion-change threshold & 2.0 original-image pixels \\
Image-difference threshold & 0.015 (normalized intensity) \\
\bottomrule
\end{tabular}
\end{table}

Each fingertip's score is divided by the threshold for its active measure.
The larger bilateral score determines the peak transition. Ties select the
first maximum in chronological order. Endpoints are always retained; the
two samples immediately before and after the peak transition are added only
when its normalized score reaches one. Indices are shared across fingers, sorted and deduplicated.
A single-sample segment retains its sole sample.

\subsection{Tactile representation and input packaging}
\label{app:tactile-representation}
\label{app:expert-contact-evolution}

Figure~\ref{fig:app-tactile-representation} shows a qualitative example
from an Insert Tube expert demonstration. Selected retains the segment
endpoints at recorded steps 598 and 778 and the adjacent intermediate
observations at steps 728 and 730. The figure displays the endpoints and
step 730, with synchronized wrist-camera and right-fingertip observations.
Green annotations mark the same tactile image region in each column.
The wider blue band in the intermediate frame illustrates information
omitted by endpoint-only sampling.

\section{Behavioral Evidence and Case Studies}
\label{app:agent-decisions}

We present selected episodes to connect demonstration content
with observable agent decisions. Each example relates the
returned observations, the agent's public messages, its subsequent
tool calls, and the resulting execution. Quotations report the
agent's interpretation; executed movements and native task
outcomes are recorded separately.

\paragraph{Behavioral counting.}
\label{app:behavior-counting}

The behavioral counts in Section~\ref{sec:experiments} are computed by
scripts from the locked episode records, rather than from the agent's
self-reported intentions. Each denominator includes all 100 query episodes
for the stated task and condition, irrespective of success. Execution
events are host tool-trace entries containing a resolved requested target
and execution feedback; previews are excluded.

Table~\ref{tab:case-index} summarizes the analyzed cases.

\begin{table}[H]
\centering\small
\caption{Cases included in the qualitative analyses.}
\label{tab:case-index}
\begin{tabular}{lll}
\toprule
Case & Task & Setting \\
\midrule
Target arrival without success & Lift Bottle & Tactile--motion \\
Uprighting (success) & Lift Bottle & Selected \\
Uprighting (failure) & Lift Bottle & Endpoints \\
Supplementary success & Lift Bottle & Endpoints \\
Supplementary failure & Lift Bottle & Selected \\
Contact adjustment & Insert Tube & Visual--motion \\
Contact adjustment & Insert Tube & Tactile-ETA \\
Stage transition (success) & Pull Out Key & Tactile--motion \\
Stage transition (failure) & Pull Out Key & Motion records \\
\bottomrule
\end{tabular}
\end{table}

\subsection{Initial grasp and object outcomes on Lift Bottle}
\label{app:bottle-decisions}

We examine the first execution event. It qualifies when its requested
gripper command is \texttt{close}, the norm of its resolved target
translation relative to the pre-execution TCP is at most 1 mm, and its
requested orientation change is at most $1^\circ$. An episode with no
execution event does not qualify. This measures the initial requested
strategy, not whether the fingers actually closed or a stable grasp was
achieved. The counts are 1/100 without demonstrations (with current tactile
input), 100/100 with visual--motion demonstrations, and 100/100 with
Tactile-ETA.

The tactile--motion failure case illustrates the
difference between reaching a robot target and completing object
manipulation. All 24 executed motion requests report target
arrival, yet the task remains unsuccessful. The recorded views
show the bottle still inclined after partial reorientation.
Opening and lowering the gripper, followed by another closure,
leaves the bottle back on the surface; subsequent regrasp attempts
do not complete the task.

This episode distinguishes controller target arrival from successful
object manipulation: reaching the requested end-effector poses does not
ensure that the resulting object pose and grasp support the next stage.

\paragraph{Uprighting and recovery in executed episodes.}
\label{app:uprighting-execution}

The successful Selected episode and the failed Endpoints episode
both begin with grasping, lifting, and an initial
orientation adjustment. Their next movements differ.
The Selected agent maintains its wrist orientation and continues
with translations that bring the bottle upright.
The Endpoints agent requests additional rotation together with
translation. This movement misses its position target, the bottle
falls back onto the surface, and subsequent recovery remains
unsuccessful.

A successful initial grasp does not determine the outcome of later
reorientation. These episodes have different initial conditions and illustrate
execution patterns rather than a matched causal comparison.

An additional successful Endpoints episode also continues
with translations after an orientation adjustment. A Selected
failure case instead loses the object during further
reorientation. These counterexamples show that the representation does not prescribe a
universally successful action sequence.

\subsection{Contact-guided adjustment on Insert Tube}
\label{app:tube-decisions}

For each executed segment, we project the measured net TCP displacement
onto that episode's initial \texttt{approach\_world} axis. We count an
episode if it fails with \texttt{native\_early\_stop}, contains at least two
segments with projected displacement greater than 1 mm, and contains no
segment below $-1$ mm. The forward segments need not be adjacent; all
executed segments are considered. Displacements exactly at the thresholds
do not qualify as forward or retreating. This is a segment-level measure,
not a claim that no backward motion occurred within any segment. The counts
are 48/100 with visual--motion demonstrations and 13/100 with Tactile-ETA.

Figure~\ref{fig:rollouts} contrasts visual--motion and
tactile-augmented demonstrations for the same initial condition.
Both conditions retain current vision, tactile observations, and proprioception.
In the visual--motion example, the agent continues advancing
until unsuccessful early termination. In the tactile-augmented
example, it adjusts its pose and subsequently retreats and shifts
laterally before completing the task.

After receiving the tactile observations following action 7,
the tactile-augmented agent states:
\begin{agentmessage}{Tactile-ETA}{Insert Tube / after action 7}
``The left tactile image now shows pronounced distortion.
I'll back off slightly and make a small lateral correction
before advancing again.''
\end{agentmessage}
The next motion request specifies a world-frame displacement
of $[-2,-2,3.46]$\,mm. Its recorded execution includes the retreat
and lateral correction illustrated in Figure~\ref{fig:rollouts}.
The agent subsequently resumes advancing and reaches native
success. This sequence shows the link between the stated contact
observation and the corrective movement actually executed.

\subsection{Stage transitions on Pull Out Key}
\label{app:key-decisions}

We examine a successful episode with tactile--motion demonstrations
and a failed episode with numerical motion records alone. Both agents
receive current visual and tactile observations and proprioception.
Their expert demonstrations use
the same two expert episodes and follow a rotate--reverse--lift
procedure. The difference lies in when the agents decide to
transition from rotation to pulling.

After its third forward rotation, the agent with tactile--motion
demonstrations states:
\begin{agentmessage}{Tactile--motion}{Pull Out Key / after the third forward rotation}
``The key remains held between both tactile pads, and the rotation
has progressed without a clear stop yet. I'll continue in smaller
increments while keeping the TCP centered over the slot.''
\end{agentmessage}
It then requests and executes a fourth forward rotation. After
receiving the next observation, it states:
\begin{agentmessage}{Tactile--motion}{Pull Out Key / after the fourth forward rotation}
``The latest turn produced a pronounced change in the left contact
patch, consistent with loading against the rotational stop. I'll
ease the rotation back slightly, then pull vertically while
preserving the grasp.''
\end{agentmessage}
Its subsequent tool calls reverse the rotation and request a
30\,mm upward motion. The episode ends in native success.

The agent given numerical motion records alone also describes
the current tactile images. After its third forward rotation,
it states:
\begin{agentmessage}{Motion records}{Pull Out Key / after the third forward rotation}
``Both tactile images show pronounced deformation during the turn,
consistent with loading against resistance. I'll ease the rotation
back slightly, then pull vertically while keeping the grasp.''
\end{agentmessage}
It then reverses the rotation and requests a 30\,mm upward motion.
This episode ends in failure.

The cases differ in when rotation gives way to pulling, despite both
agents receiving current tactile observations. They illustrate stage-transition decisions;
the aggregate comparison is reported in Table~\ref{tab:ablation-designs}.

\begin{figure}[H]
\centering
\includegraphics[width=\linewidth]{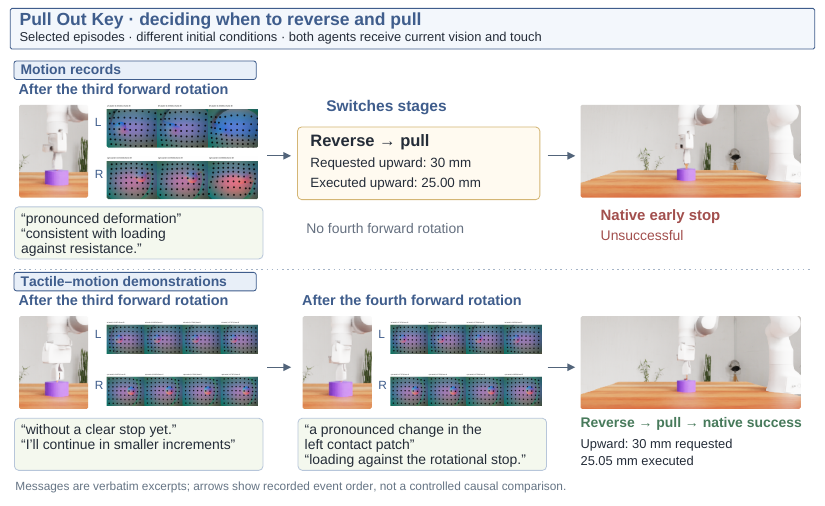}
\caption{Stage transitions in two Pull Out Key episodes. The motion-record
agent reverses after three forward rotations; the tactile--motion agent
performs a fourth before reversing and pulling. Images and message excerpts
show the recorded sequence. Initial conditions differ; outcomes are native labels.}
\label{fig:app-demonstration-content}
\end{figure}

\section{Effect of Demonstration Count}
\label{app:additional-ablations}

We vary the number of complete expert episodes supplied to Tactile-ETA,
using GPT-6 Astra with \texttt{low} reasoning effort and the same four tasks
and 100 initial conditions per task as the main comparison. Each demonstration
includes visual observations, bilateral tactile observations, and numerical
records. The 2-shot condition uses expert episodes 0 and 1; 4-shot adds the
next two successful official episodes in ID order. For 1-shot, even-indexed
queries use episode 0 and odd-indexed queries use episode 1, with 50 queries
each. Current sensing and the execution interface are unchanged.

\begin{table}[H]
\centering\small
\caption{Success rate (\%) with 1, 2, or 4 complete demonstrations.
Each task--setting pair contains 100 evaluable query episodes.
The shaded column denotes the 2-shot setting used in the main comparisons.}
\label{tab:shot-scaling}
\setlength{\tabcolsep}{12pt}
\begin{tabular}{l r >{\columncolor{appendixblue}}r r}
\toprule
Task & 1-shot & 2-shot & 4-shot \\
\midrule
Insert Tube & 22 & \textbf{32} & 23 \\
Lift Can & 2 & 2 & \textbf{3} \\
Lift Bottle & 31 & \textbf{36} & 2 \\
Pull Out Key & 35 & 26 & \textbf{48} \\
\midrule
Average & 22.50 & \textbf{24.00} & 19.00 \\
\bottomrule
\end{tabular}
\end{table}

More demonstrations do not consistently improve performance
(Table~\ref{tab:shot-scaling}). Increasing from two to four demonstrations
improves Pull Out Key but reduces success on Insert Tube and Lift Bottle.

\paragraph{Trial accounting.}
The first pass produced 1,086 evaluable outcomes. A separate supplement
completed 114 missing query--setting combinations after initialization
unavailability (104), infrastructure issues (6), or cancellation (4).
The supplement removed initialization and worker-startup deadlines while
retaining operation and cleanup budgets. Existing valid task failures were
not rerun. All rates use the resulting 100 evaluable outcomes per task and
shot count, combining the two initialization policies. Supplement counts
for 1/2/4-shot are 7/6/6 on Insert Tube, 26/31/18 on Lift Can, 5/3/1 on
Lift Bottle, and 2/4/5 on Pull Out Key.

\Needspace{12\baselineskip}
\section{Real-World Experiments}
\label{app:realworld-details}

\paragraph{Robot platform.}
Both tasks use a reBot Arm B601-RS with a vitai GF225 tactile gripper.

\paragraph{Agent and demonstrations.}
Both tasks use GPT-6 Astra with \texttt{low} reasoning effort.
The no-demonstration and Tactile-ETA configurations both receive current
visual and tactile observations and robot state.
Tactile-ETA additionally receives one manually recorded demonstration per
task containing visual observations, robot motion, and tactile observations.
The demonstration is supplied in context without updating model parameters.
Each task is evaluated with five trials per configuration.

\paragraph{Success criteria.}
Task success is judged by a human evaluator.
For Wipe Board, success requires removing the target blue writing from
the whiteboard, with the eraser moved away so that the cleaned region is
visible. For Pick Up Chips, success requires lifting the potato chip fully
clear of its support and holding it without visible breakage or dropping it;
the support and its base must remain on the table.
Figure~\ref{fig:realworld}(b) reports the task completion counts.

\Needspace{10\baselineskip}
\section{Prompts and Tool Interface}
\label{app:tools-and-prompts}

\subsection{Operator prompts}
\label{app:operator-prompts}
The three shared blocks below replace only the task name, goal, static rules,
and control-step limit with named fields; task-specific blocks provide their
values. Tool argument schemas follow in Section~\ref{app:argument-schemas}.

\begin{appendixprompt}{Shared operator prompt: interaction}
\raggedright
You control one live UniVTAC \{task\_name\} episode through OpenETA's general-purpose robot tools.
\par\smallskip
Start by calling review\_demonstrations once, then observe the current episode. Use the current camera images, bilateral tactile images when provided, robot state, and execution feedback to decide your motion targets, orientation, and gripper actions.
\par\smallskip
After each action, inspect the returned observations. You may probe, retreat, adjust, and retry within this episode. A recoverable tool failure does not end the task.
\par\smallskip
Use finish\_episode to end voluntarily or acknowledge the neutral episode-ended notification. Voluntary ending is final. Native evaluation and automatic stopping run in the background; current task scores and the specific termination reason are not provided. The demonstration tool may provide historical expert experience or no examples. Use any examples as references for reasoning about the current observations, not as absolute-coordinate scripts. Recorded measured movement is not necessarily a recorded tool command. Historical success does not guarantee current success.
\par\smallskip
Do not infer task success merely because a command completed.
\end{appendixprompt}

\begin{appendixprompt}{Shared operator prompt: coordinates and observations}
\raggedright
Task goal: \{task\_goal\}
\par\smallskip
Coordinates: xyz\_m is the absolute TCP position in world metres; delta\_mm is displacement in world or current grip\_site axes, in millimetres. The TCP is UniVTAC's measured gripper center (fixed transform from panda\_hand), not the LIBERO grip site. approach\_world sets TCP +Z; jaw\_world sets TCP +X. These fields specify orientation axes, not a task-specific rotation axis.
\par\smallskip
Unspecified position/orientation/gripper remain unchanged. An omitted gripper preserves the previous two finger actuator targets, including the commands inherited from reset. Explicit open/close changes and retains them. Arm arrival and completion of a finite gripper wait are separate feedback; neither establishes a stable grasp or task success.
\par\smallskip
After move\_to, head/wrist show the current scene and the two tactile image strips show, when provided, real segment-end tactile history on a shared time axis, at up to four times. These are selected image/motion changes, not slip labels or early warnings. The takeover reference is not necessarily a no-contact baseline.
\par\smallskip
Use preview=true for target resolution without physics. Use execute\_preview\_id alone to execute that preview. A preview is valid until the next real request.
\end{appendixprompt}

\begin{appendixprompt}{Shared operator prompt: budgets and task requirements}
\raggedright
Shared episode budgets and termination semantics: At most 30 admitted non-preview move\_to requests, 100 MCP calls and 3600 seconds of total Codex wall time are available. Planning rejection of an admitted motion request consumes its request budget. Native control is limited to \{native\_control\_step\_limit\} steps, with at most 80 control steps per motion segment. Each control step advances two physics steps at 120 Hz; physics pauses while you reason. A segment limit need not end the episode: inspect the actual feedback. Native terminal or an episode budget ends physical interaction. After native terminal, last-observation and completion tools remain available for up to 300 seconds, within the total Codex wall-time limit. Finish naturally when done.
\par\smallskip
Static native rules: \{static\_native\_rules\}
\par\smallskip
Some observation modalities may be absent. Use only the images and measurements actually returned. Missing tactile input is not a tool failure.
\end{appendixprompt}

\begin{appendixprompt}{Insert Tube task fields}
\raggedright
task\_name: insert tube
\par\smallskip
native\_control\_step\_limit: 300
\par\smallskip
Task goal: Insert the 2.0 cm diameter test tube into the 2.05 cm narrow hole on the inclined surface from the near-center starting position.
\par\smallskip
Static native rules: the object's pose origin in the hole frame must satisfy x\ensuremath{<}0.005 m, y\ensuremath{<}0.005 m (these are signed comparisons, not absolute values), z\ensuremath{<}-0.03 m, and the object's positive z axis dotted with the hole positive z axis \ensuremath{>}0.965. The absolute change of object z relative to the gripper from the native reference must be \ensuremath{<}0.03 m. That change \ensuremath{>}0.03 m triggers early termination. Equality does not satisfy success or that early-stop test. Object axial position is not TCP displacement. Current hole transform and grip-relative drift are not supplied. These are static requirements, not current measurements. Only neutral episode-ended feedback is returned.
\end{appendixprompt}

\begin{appendixprompt}{Lift Can task fields}
\raggedright
task\_name: lift can
\par\smallskip
native\_control\_step\_limit: 300
\par\smallskip
Task goal: Grasp the horizontal cylindrical can on the ground plane and lift it vertically without slippage.
\par\smallskip
Static native rules: can world pose-origin z \ensuremath{<}0.01 m and absolute dot product of its x axis with world [0,0,1] \ensuremath{>}0.99. With the configured gsmini sensor, maximum native indentation depth \ensuremath{>}8.0 mm causes early termination. Also, absolute in-hand z change from the native reference \ensuremath{>}0.05 m together with absolute dot product of the can z axis with world [0,0,1] \ensuremath{>}0.99 causes early termination. These are the native predicates, not a minimum lift-height rule. Current indentation, in-hand drift and object ground truth are not supplied. These are static requirements, not current measurements. Only neutral episode-ended feedback is returned.
\end{appendixprompt}

\begin{appendixprompt}{Lift Bottle task fields}
\raggedright
task\_name: lift bottle
\par\smallskip
native\_control\_step\_limit: 500
\par\smallskip
Task goal: Grasp the bottle resting on the ground plane and lift it vertically, keeping the bottle base within 5 cm of the wall at the final pose.
\par\smallskip
Static native rules: bottle origin in the native target frame must satisfy x\ensuremath{>}-0.02 m, \ensuremath{\vert}y\ensuremath{\vert}\ensuremath{<}0.1 m and \ensuremath{\vert}z\ensuremath{\vert}\ensuremath{<}0.001 m; the absolute dot product of its x axis expressed in that frame with [0,0,1] must be \ensuremath{>}0.99. After more than 300 native action steps, absolute dot product of that x axis with [-1,0,0] \ensuremath{>}0.99 causes early termination. These are target-relative pose comparisons, not a minimum lift-height rule. The current target transform is not supplied. These are static requirements, not current measurements. Only neutral episode-ended feedback is returned.
\end{appendixprompt}

\begin{appendixprompt}{Pull Out Key task fields}
\raggedright
task\_name: pull out key
\par\smallskip
native\_control\_step\_limit: 300
\par\smallskip
Task goal: Rotate the key in the slot until mechanical resistance is sensed, then pull the key straight out.
\par\smallskip
Static native rules: key world z \ensuremath{>}0.09 m, key positive z axis dotted with world positive z \ensuremath{>}0.965, and slot x axis expressed relative to its native initial reference dotted with [1,0,0] \ensuremath{>}0.99. The absolute difference between native robot-base-relative end-effector z and key world z must be \ensuremath{<}0.14 m. That difference \ensuremath{\geq}0.14 m or the slot-axis dot product \ensuremath{<}0.99 triggers early termination. These are the native comparisons; current key/slot ground truth and these measured values are not supplied. These are static requirements, not current measurements. Only neutral episode-ended feedback is returned.
\end{appendixprompt}

\subsection{Tool interface}
Table~\ref{tab:robot-interface} summarizes the exposed tools. The online TCP
is the measured gripper center, offset by 0.131 m along the local hand axis
from \texttt{panda\_hand}; online state quaternions use \texttt{xyzw} order.
The argument conventions below therefore differ from the historical
base-relative hand poses. Omitted pose components and gripper targets remain
unchanged. A preview resolves a target without advancing physics; executing
it uses its returned identifier, which expires on the next real request.

\begin{table}[H]
\centering\small
\caption{Robot interface. Auxiliary tools expose no online native task score.}
\label{tab:robot-interface}
\begin{tabular}{p{0.30\linewidth}p{0.62\linewidth}}
\toprule
Tool & Arguments and purpose \\
\midrule
\texttt{review\_demonstrations} & No arguments; retrieve demonstration context. \\
\texttt{observe} & No arguments; obtain current observations. \\
\texttt{move\_to} & Absolute \texttt{xyz\_m} in world metres or relative
\texttt{delta\_mm} in world/current \texttt{grip\_site} axes;
\texttt{approach\_world} and \texttt{jaw\_world} specify TCP $+Z$ and $+X$;
optional gripper, preview, and preview identifier. \\
\texttt{mark\_point} & \texttt{view}, integer pixel \texttt{u,v}, optional
\texttt{point\_id}; identify an image point. \\
\texttt{report\_issue} & Required \texttt{message}; report an issue. \\
\texttt{finish\_episode} & Optional \texttt{reason}; end voluntarily or
acknowledge termination. \\
\bottomrule
\end{tabular}
\end{table}

\subsection{Argument schemas}
\label{app:argument-schemas}
The core motion schema below reproduces the retained argument definition,
omitting only display-only \texttt{title} fields. Auxiliary tool arguments
are summarized in Table~\ref{tab:robot-interface}. The protocol declares
\texttt{structured\_output=false}. Array lengths are not constrained by these
schemas; the three-axis interpretation is specified in the operator prompt.

\begin{appendixprompt}{move\_to argument schema}
\raggedright
\ttfamily
\{
\par\smallskip
"properties": \{
\par\smallskip
"xyz\_m": \{"anyOf": [\{"items": \{"type": "number"\}, "type": "array"\}, \{"type": "null"\}], "default": null\},
\par\smallskip
"delta\_mm": \{"anyOf": [\{"items": \{"type": "number"\}, "type": "array"\}, \{"type": "null"\}], "default": null\},
\par\smallskip
"delta\_frame": \{"default": "world", "enum": ["world", "grip\_site"], "type": "string"\},
\par\smallskip
"approach\_world": \{"anyOf": [\{"items": \{"type": "number"\}, "type": "array"\}, \{"type": "null"\}], "default": null\},
\par\smallskip
"jaw\_world": \{"anyOf": [\{"items": \{"type": "number"\}, "type": "array"\}, \{"type": "null"\}], "default": null\},
\par\smallskip
"gripper": \{"anyOf": [\{"enum": ["open", "close"], "type": "string"\}, \{"type": "null"\}], "default": null\},
\par\smallskip
"preview": \{"default": false, "type": "boolean"\},
\par\smallskip
"execute\_preview\_id": \{"anyOf": [\{"type": "string"\}, \{"type": "null"\}], "default": null\}
\par\smallskip
\},
\par\smallskip
"type": "object"
\par\smallskip
\}
\end{appendixprompt}

\end{document}